# From influence diagrams to multi-operator cluster DAGs

**Cédric Pralet**
LAAS-CNRS
Toulouse, France

**Thomas Schiex**
INRA
Castanet Tolosan, France

**Gérard Verfaillie**
ONERA
Toulouse, France

## Abstract

There exist several architectures to solve influence diagrams using local computations, such as the Shenoy-Shafer, the HUGIN, or the Lazy Propagation architectures. They all extend usual variable elimination algorithms thanks to the use of so-called "potentials". In this paper, we introduce a new architecture, called the Multi-operator Cluster DAG architecture, which can produce decompositions with an improved *constrained induced-width*, and therefore induce potentially exponential gains. Its principle is to benefit from the composite nature of influence diagrams, instead of using uniform potentials, in order to better analyze the problem structure.

## 1 INTRODUCTION

Since the first algorithms based on decision trees or arc-reversal operations [Shachter, 1986], several exact methods have been proposed to solve influence diagrams using local computations, such as the ones based on the Shenoy-Shafer, the HUGIN, or the Lazy Propagation architectures [Shenoy, 1992; Jensen *et al.*, 1994; Madsen and Jensen, 1999]. These methods have successfully adapted classical Variable Elimination (VE) techniques (which are basically designed to compute one type of marginalization on a combination of local functions with only one type of combination operator), in order to handle the multiple types of information (probabilities and utilities), the multiple types of marginalizations (sum and max), and the multiple types of combination ($\times$ for probabilities, $+$ for utilities) involved in an influence diagram. The key mechanism used for such an extension consists in using elements known as *potentials* [Ndilikilikesha, 1994].

In this paper, we define a new architecture, called the Multi-operator Cluster DAG (MCDAG) architecture, which does not use potentials, but still relies on VE. Compared to existing schemes, MCDAGs actively exploit the composite nature of influence diagrams. We first present the potential-based approach and motivate the need for a new architecture (Section 2). Then, MCDAGs are introduced (Section 3) and a VE algorithm is defined (Section 4). Finally, this work is compared with existing approaches (Section 5) and extended to other frameworks (Section 6). All proofs are available in [Pralet *et al.*, 2006b].

## 2 MOTIVATIONS

**Notations and definitions** An *influence diagram* [Howard and Matheson, 1984] is a composite graphical model defined on three sets of variables organized in a Directed Acyclic Graph (DAG) $G$: (1) a set $C$ of *chance variables* $x \in C$, for each of which a conditional probability distribution $P_{x \mid pa(x)}$ on $x$ given its parents in $G$ is specified; (2) a set $D = \{D_1, \ldots, D_q\}$ (indices represent the order in which decisions are made) of *decision variables* $x \in D$, for each of which $pa(x)$ is the set of variables observed before decision $x$ is made; (3) a set $\Gamma$ of *utility variables* $u \in \Gamma$, each of which is associated with a utility function $U_{pa(u)}$ on $pa(u)$ (and utility variables are leaves in the DAG).

We consider influence diagrams where the parents of a decision variable are parents of all subsequent decision variables (no-forgetting). The set of conditional probability distributions (one for each $x \in C$) is denoted $P$ and the set of utility functions (one for each $u \in \Gamma$) is denoted $U$. Each function $\phi \in P \cup U$ holds on a set of variables $sc(\phi)$ called its scope, and is consequently called a *scoped function* ($sc(P_{x \mid pa(x)}) = \{x\} \cup pa(x)$ and $sc(U_{pa(u)}) = pa(u)$). The set of chance variables observed before the first decision is denoted $I_0$, the set of chance variables observed between decisions $D_k$ and $D_{k+1}$ is denoted $I_k$, and the set of chance variables unobserved before the last decision is denoted $I_q$. We use $dom(x)$ to denote the domain of a vari-able $x \in C \cup D$, and by extension, for $W \subset C \cup D$, $dom(W) = \prod_{x \in W} dom(x)$.

The usual problem associated with an influence diagram is to find *decision rules* maximizing the expected utility (a decision rule for a decision $D_k$ is a function associating a value in $dom(D_k)$ with any assignment of the variables observed before making decision $D_k$) As shown in [Jensen *et al.*, 1994], this is equivalent to computing optimal decision rules for the quantity

$$\sum_{I_0} \max_{D_1} \ldots \sum_{I_{q-1}} \max_{D_q} \sum_{I_q} \left( \left( \prod_{P_i \in P} P_i \right) \times \left( \sum_{U_i \in U} U_i \right) \right) \quad (1)$$

## 2.1 THE "POTENTIAL" APPROACH

With this approach, Equation 1 is reformulated using so-called *potentials* in order to use only one combination and one marginalization operator. A potential on a set of variables $W$ is a pair $\pi_W = (p_W, u_W)$, where $p_W$ and $u_W$ are respectively a nonnegative real function and a real function, whose scopes are included in $W$. The initial conditional probability distributions $P_i \in P$ are transformed into potentials $(P_i, 0)$, whereas the initial utility functions $U_i \in U$ are transformed into potentials $(1, U_i)$. On these potentials, a *combination* operation $\otimes$ and a *marginalization* (or *elimination*) operation $\uparrow$ are defined:

- the combination of $\pi_{W_1} = (p_{W_1}, u_{W_1})$ and $\pi_{W_2} = (p_{W_2}, u_{W_2})$ is the potential on $W_1 \cup W_2$ given by $\pi_{W_1} \otimes \pi_{W_2} = (p_{W_1} \times p_{W_2}, u_{W_1} + u_{W_2})$;

- the marginalization of $\pi_W = (p_W, u_W)$ over $W_1 \subset C$ equals $\pi_W^{\uparrow W_1} = \left( \sum_{W_1} p_W, \frac{\sum_{W_1} p_W u_W}{\sum_{W_1} p_W} \right)$ (with the convention $0/0 = 0$), whereas the marginalization of $\pi_W = (p_W, u_W)$ over $W_1 \subset D$ is given by $\pi_W^{\uparrow W_1} = (p_W, \max_{W_1} u_W)$.

Solving the problem associated with an influence diagram is then equivalent to computing $\beta = ((\cdots((\pi_{C \cup D}{}^{\uparrow I_q})^{\uparrow D_q})^{\uparrow I_{q-1}} \cdots)^{\uparrow D_1})^{\uparrow I_0}$, where $\pi_{C \cup D} = (\otimes_{P_i \in P} (P_i, 0)) \otimes (\otimes_{U_i \in U} (1, U_i))$ is the combination of the initial potentials. As $\otimes$ and $\uparrow$ satisfy the Shenoy-Shafer axioms defined in [Shenoy, 1990], $\beta$ can be computed using usual VE algorithms [Jensen *et al.*, 1994]. This explains why existing architectures like Shenoy-Shafer, HUGIN, or Lazy Propagation (LP[1]) use potentials to solve influence diagrams.

## 2.2 QUANTIFYING THE COMPLEXITY

In the case of influence diagrams, the alternation of sum and max marginalizations, which do not gener-

---

[1]The LP architecture actually uses potentials defined as pairs of set of functions (instead of pairs of functions).

---

ally commute, prevents from eliminating variables in any order. The complexity of VE can then be quantified using *constrained induced-width* [Jensen *et al.*, 1994; Park and Darwiche, 2004] (instead of induced-width [Dechter and Fattah, 2001]).

**Definition 1.** *Let $G = (V_G, H_G)$ be a hypergraph[2] and let $\preceq$ be a partial order on $V_G$. The constrained induced-width of $G$ with constraints on the elimination order given by $\preceq$ ("$x \prec y$" stands for "$y$ must be eliminated before $x$") is a parameter denoted $w_G(\preceq)$. It is defined as $w_G(\preceq) = \min_{o \in lin(\preceq)} w_G(o)$, $lin(\preceq)$ being the set of linearizations of $\preceq$ to a total order on $V_G$ and $w_G(o)$ being the induced-width of $G$ for the elimination order $o$ (i.e. the size of the largest hyperedge created when eliminating variables in the order given by $o$).*

The constrained induced-width can be used to give an upper bound on the complexity of existing potential-based VE algorithms. Let $G_p = (C \cup D, \{sc(\phi) | \phi \in P \cup U\})$ be the hypergraph corresponding to the "untyped" influence diagram. Let $\preceq_p$ be the partial order defined by $I_0 \prec_p D_1$, $(I_k \neq \emptyset) \rightarrow (D_k \prec_p I_k \prec_p D_{k+1})$, and $D_q \prec_p I_q$. Finally, let $d$ be the maximum size of the variables domains. Then, with classical approaches based on potentials and *strong junction trees* [Jensen *et al.*, 1994], which are junction trees with constraints on the marginalization order, the theoretical complexity is $O(|P \cup U| \cdot d^{1+w_{G_p}(\preceq_p)})$ (the number of elements of a finite set $E$ is denoted $|E|$).

## 2.3 DECREASING THE CONSTRAINED INDUCED-WIDTH

The constrained-induced width is a guideline to show how the complexity can be decreased. In this direction, one can work on the two parameters on which it depends: the partial order $\preceq$, and the hypergraph $G$.

**Weakening the partial order $\preceq$**

**Proposition 1.** *Let $G = (V_G, H_G)$ be a hypergraph and let $\preceq_1, \preceq_2$ be two partial orders on $V_G$ such that $\forall (x, y) \in V_G \times V_G, (x \preceq_2 y) \rightarrow (x \preceq_1 y)$ ($\preceq_2$ is weaker than $\preceq_1$). Then, $w_G(\preceq_1) \geq w_G(\preceq_2)$.*

Proposition 1 means that if one weakens $\preceq$, i.e. if one reveals some extra freedoms in the elimination order (e.g. by proving that some marginalizations with sum and max can commute), then the theoretical complexity may decrease. Though such a technique is known to be useless in contexts like Maximum A Posteriori hypothesis [Park and Darwiche, 2004], where there is only one alternation of max and sum marginalizations,

---

[2]This means that $V_G$ is the set of variables (or vertices), and $H_G$ is a set of hyperedges on $V_G$, i.e. a subset of $2^{V_G}$.

it can lead to an exponential gain as soon as there are more than two levels of alternation.

Indeed, assume that one wants to compute $\max_{x_1,\ldots,x_n} \sum_y \max_{x_{n+1}} P_y(U_{x_1,y} + \sum_{1 \leq i \leq n} U_{x_i,x_{n+1}})$. On one hand, using $\preceq_1$ defined by $\{x_1,\ldots,x_n\} \prec_1 y \prec_1 x_{n+1}$ provides us with the constrained induced-width $w_G(\preceq_1) = n$, since $x_{n+1}$ is then necessarily eliminated first. On the other hand, the scopes of the functions involved enable us to infer that with $\preceq_2$ defined by $x_1 \prec_2 y$, one is guaranteed to compute the same value, since $y$ is "linked" only with $x_1$. The constrained induced-width is then $w_G(\preceq_2) = 1$, e.g. with the elimination order $x_1 \prec y \prec x_{n+1} \prec x_n \prec \ldots \prec x_2$. Therefore, the theoretical complexity decreases from $O((n+2) \cdot d^{n+1})$ to $O((n+2) \cdot d^2)$, thanks to the weakening of the partial order (the $(n+2)$ factor corresponds to the number of scoped functions).

**Working on the hypergraph** The second possible mechanism is to work on the hypergraph $G$, either by eliminating so-called "barren" variables (computing $\sum_x P_{x|pa(x)}$ is useless because of normalization), or by better decomposing the problem. To illustrate the latter, assume that one wants to compute $\max_{x_1,\ldots,x_n} \sum_y P_y \cdot (U_{y,x_1} + \cdots + U_{y,x_n})$. The basic hypergraph $G_1 = (\{x_1,\ldots,x_n,y\},\{\{y,x_1\},\ldots,\{y,x_n\}\})$, together with $\preceq_1$ defined by $\{x_1,\ldots,x_n\} \prec_1 y$, gives a theoretical complexity $O((n+1) \cdot d^{w_{G_1}(\preceq_1)+1}) = O((n+1) \cdot d^{n+1})$. However, one can write:

$\max_{x_1,\ldots,x_n} \sum_y P_y \cdot (U_{y,x_1} + \cdots + U_{y,x_n})$
$= (\max_{x_1} \sum_y P_y \cdot U_{y,x_1}) + \cdots + (\max_{x_n} \sum_y P_y \cdot U_{y,x_n})$

Thus, an implicit *duplication* of $y$ makes the complexity decrease to $O((n+1)d^2) = O((n+1)d^{1+w_{G_2}(\preceq_2)})$, where $G_2$ is the hypergraph defined by the variables $\{x_1,\ldots,x_n,y^{(1)},\ldots,y^{(n)}\}$ and by the hyperedges $\{\{x_1,y^{(1)}\},\ldots,\{x_n,y^{(n)}\}\}$, and where $\preceq_2$ is given by $x_1 \prec_2 y^{(1)}, \ldots, x_n \prec_2 y^{(n)}$. This method, which uses the property $\sum_S (U_1 + U_2) = (\sum_S U_1) + (\sum_S U_2)$, duplicates variables "quantified" with $\sum$, so that computations become more local. Proposition 2 shows the possible exponential gain obtained by duplication.

**Proposition 2.** *Let $\phi_{x,S_i}$ be a scoped function of scope $\{x\} \cup S_i$ for any $i \in [1,m]$. The direct computation of $\sum_x (\phi_{x,S_1} + \cdots + \phi_{x,S_m})$ always requires more sums than the direct computation of $(\sum_x \phi_{x,S_1}) + \cdots + (\sum_x \phi_{x,S_m})$. Moreover, the computation of $\sum_x (\phi_{x,S_1} + \cdots + \phi_{x,S_m})$ results in a complexity $O(m \cdot d^{1+|S_1 \cup \ldots \cup S_m|})$, whereas the computation of the $m$ quantities in the set $\{\sum_x \phi_{x,S_i} \mid 1 \leq i \leq m\}$ results in a complexity $O(m \cdot d^{1+\max_{i \in [1,m]} |S_i|})$.*

**Why not use potentials?** Though weakening the constraints on the elimination order could be done with potentials, the duplication mechanism cannot be used if potentials are. Indeed, one cannot write $(\pi_{W_1} \otimes \pi_{W_2})^{\uparrow W_3} = (\pi_{W_1}^{\uparrow W_3}) \otimes (\pi_{W_2}^{\uparrow W_3})$ even if $W_3 \subset C$. The duplication mechanism has actually already been proposed in the influence diagram litterature [Dechter, 2000] where it was applied "on the fly" during elimination. In this paper, the duplication is exploited in a global preliminary analysis which may reveal new degrees of freedom in the elimination order, in synergism with the application of other mechanisms. The new architecture we introduce, which does not use potentials to solve influence diagrams, is called the *Multi-operator Cluster DAG* (MCDAG) architecture.

## 3 THE MCDAG ARCHITECTURE

### 3.1 MACROSTRUCTURING AN INFLUENCE DIAGRAM

The first step to build the MCDAG architecture is to analyze the *macrostructure* of the influence diagram, by detecting the possible reordering freedoms in the elimination order, while using the duplication technique and the normalization conditions on conditional probability distributions. This macrostructure is represented with a DAG of *computation nodes*.

**Definition 2.** *An atomic computation node $n$ is a scoped function $\phi$ in $P \cup U$. In this case, the value of $n$ is $val(n) = \phi$, and its scope is $sc(n) = sc(\phi)$. A computation node is either an atomic computation node or a triple $n = (Sov, \circledast, N)$, where $Sov$ is a sequence of operator-variables pairs, $\circledast$ is an associative and commutative operator with an identity, and where $N$ is a set of computation nodes. In the latter, the value of $n$ is given by $val(n) = Sov(\circledast_{n' \in N} val(n'))$, and its scope is given by $sc(n) = (\cup_{n' \in N} sc(n')) - \{x \mid op_x \in Sov\}$.*

Informally, a computation node $(Sov, \circledast, N)$ defines a sequence of marginalizations on a combination of computation nodes with a specific operator. It can be represented as in Figure 1. Given a set of computation nodes $N$, we define $N^{+x}$ (resp. $N^{-x}$) as the set of nodes of $N$ whose scope contains $x$ (resp. does not contain $x$): $N^{+x} = \{n \in N \mid x \in sc(n)\}$ (resp. $N^{-x} = \{n \in N \mid x \notin sc(n)\}$).

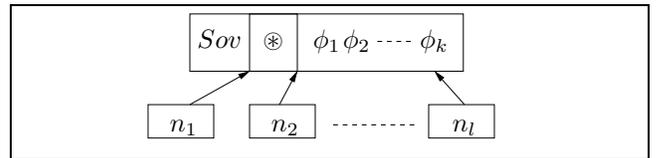

**Figure 1:** A computation node $(Sov, \circledast, N)$, where $\{\phi_1, \ldots, \phi_k\}$ (resp. $\{n_1, \ldots, n_l\}$) is the set of atomic (resp. non-atomic) computation nodes in $N$.

### 3.1.1 From influence diagrams to computation nodes

Without loss of generality, we assume that $U \neq \emptyset$ (if this is not the case, one can add $U_0 = 0$ to $U$).

**Proposition 3.** *Let $Sov_0$ be the initial sequence $\sum_{I_0} \max_{D_1} \ldots \sum_{I_{q-1}} \max_{D_q} \sum_{I_q}$ of operator-variables pairs defined by the influence diagram. The value of Equation 1 is equal to the value of the computation node $n_0 = (Sov_0, +, \{(\emptyset, \times, P \cup \{U_i\}), U_i \in U\})$.*

For the influence diagram associated with the computation of $\max_d \sum_{r_2, r_1} P_{r_1} \cdot P_{r_2|r_1} \cdot (U_{d,r_1} + U_{d,r_2} + U_d)$, $n_0$ corresponds to the first computation node in Figure 2.

### 3.1.2 Macrostructuring the initial node

In order to exhibit the macrostructure of the influence diagram, we analyze the sequence of computations performed by $n_0$. To do so, we successively consider the eliminations in $Sov_0$ from the right to the left and use three types of *rewriting rules*, preserving nodes values, to make the macrostructure explicit: (1) *decomposition rules*, which decompose the structure using namely the duplication technique; (2) *recomposition rules*, which reveal freedoms in the elimination order; (3) *simplification rules*, which remove useless computations from the architecture, by using normalization conditions. Rewriting rules are presented first for the case of sum-marginalizations, and then for the case of max-marginalizations. A rewriting rule may be preceded by preconditions restricting its applicability.

**Rewriting rules for $\sum_x$** When a sum-marginalization must be performed, a decomposition rule $D_\Sigma$, a recomposition rule $R_\Sigma$, and two simplification rules $S_\Sigma^1$ and $S_\Sigma^2$ are used. These are illustrated in Figure 2, which corresponds to the influence diagram example introduced in 3.1.1.

$\boxed{D_\Sigma}$ $(Sov.\sum_x, +, \{(\emptyset, \times, N), N \in \mathfrak{N}\})$
$\leadsto (Sov, +, \{(\emptyset, \times, N^{-x} \cup \{(\sum_x, \times, N^{+x})\}), N \in \mathfrak{N}\})$

$\boxed{R_\Sigma}$ [ Prec.: $(S' \cap (S \cup sc(N_1)) = \emptyset) \wedge (N_1 \cap N_2 = \emptyset)$ ]
$(\sum_S, \times, N_1 \cup \{(\sum_{S'}, \times, N_2)\}) \leadsto (\sum_{S \cup S'}, \times, N_1 \cup N_2)$

$\boxed{S_\Sigma^1}$ [ Prec.: $x \notin S \cup sc(N)$ ]
$(\sum_{\{x\} \cup S}, \times, N \cup \{P_{x \,|\, pa(x)}\}) \leadsto (\sum_S, \times, N)$

$\boxed{S_\Sigma^2}$ $(\emptyset, \times, N \cup \{(\sum_\emptyset, \times, \emptyset)\}) \leadsto (\emptyset, \times, N)$

**Example** In the example of Figure 2, the first rule to be applied is the decomposition rule $D_\Sigma$, which treats the operator-variable pair $\sum_{r_1}$. Such a rule uses the duplication mechanism and the distributivity property of $\times$ over $+$. It provides us with a DAG of computation nodes. It is a DAG since common computation nodes are merged (and it is not hard to detect such nodes when applying the rules). Then, $D_\Sigma$ can be applied again for $\sum_{r_2}$. One can infer from the obtained architecture that there is no reason for $r_1$ to be eliminated before $r_2$. Using the recomposition rule $R_\Sigma$ makes this clear in the structure. Basically, $R_\Sigma$ uses the distributivity of $\times$ over $+$. Last, applying $S_\Sigma^1$ and $S_\Sigma^2$, which use the normalization of conditional probability distributions, simplifies some nodes in the architecture. In the end, no computation involves more than two variables if one eliminates $r_1$ first in the node $(\sum_{r_1, r_2}, \times, \{P_{r_1}, P_{r_2|r_1}, U_{d, r_2}\})$, whereas with a poten-tial-based approach, it would be necessary to process three variables simultaneously (since $r_1$ would be involved in the potentials $(P_{r_1}, 0), (P_{r_2|r_1}, 0), (1, U_{d,r_1})$ if eliminated first, and $r_2$ would be involved in the potentials $(P_{r_2|r_1}, 0), (1, U_{d,r_2})$ if eliminated first).

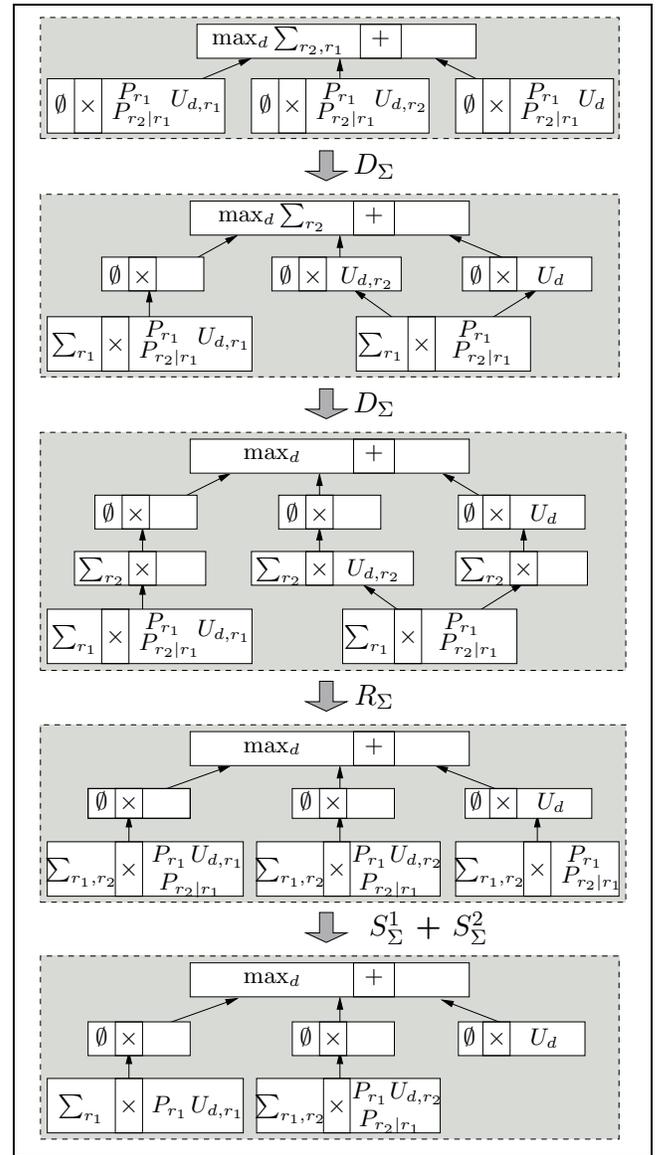

**Figure 2:** Application of rewriting rules for $\sum$.

**Rewriting rules for** $\max_x$  When a max-marginalization must be performed, a decomposition rule $D_{\max}$ and a recomposition rule $R_{\max}$ are used (there is no simplification rule since there is no normalization condition to use for decision variables). These rules are a bit more complex than the previous ones and are illustrated in Figure 3, which corresponds to the influence diagram $\max_{d_1} \sum_{r_2} \max_{d_2} \sum_{r_1} \max_{d_3} P_{r_1} \cdot P_{r_2|r_1} \cdot (U_{d_1} + U_{d_2,d_3} + U_{r_2,d_1,d_3} + U_{r_1,d_2})$.

$\boxed{D_{\max}}$ [ Prec.: $\forall N \in \mathfrak{N}^{+x} \forall n \in N^{-x}, val(n) \geq 0$ ]

$(Sov.\max_x, +, \{(\emptyset, \times, N), N \in \mathfrak{N}\})$

$\leadsto \begin{cases} (Sov, +, \{(\emptyset, \times, N), N \in \mathfrak{N}\}) \text{ if } \mathfrak{N}^{+x} = \emptyset \\ (Sov, +, \{(\emptyset, \times, N), N \in \mathfrak{N}^{-x}\} \\ \quad \cup \{(\emptyset, \times, N_1 \cup \{(\max_x, +, N_2)\})\}) \text{ otherwise} \end{cases}$

where $\begin{cases} N_1 = \cap_{N \in \mathfrak{N}^{+x}} N^{-x} \\ N_2 = \{(\emptyset, \times, N - N_1), N \in \mathfrak{N}^{+x}\} \end{cases}$

$\boxed{R_{\max}}$ [ Prec.: $(S' \cap (S \cup sc(N_1) \cup sc(N_2))) = \emptyset) \wedge (\forall N_3 \in \mathfrak{N}, N_2 \cap N_3 = \emptyset) \wedge (\forall n \in N_2, val(n) \geq 0)$ ]

$(\max_S, +, N_1 \cup$
$\quad \{(\emptyset, \times, N_2 \cup \{(\max_{S'}, +, \{(\emptyset, \times, N_3), N_3 \in \mathfrak{N}\})\})\})$
$\leadsto (\max_{S \cup S'}, +, N_1 \cup \{(\emptyset, \times, N_2 \cup N_3), N_3 \in \mathfrak{N}\})$

**Example**  In the example of Figure 3, one first applies the decomposition rule $D_{\max}$, in order to treat the operator-variable pair $\max_{d_3}$. Such a rule uses first the monotonicity of $+$ ($\max(a+b, a+c) = a + \max(b,c)$), and then both the distributivity of $\times$ over $+$ and the monotonicity of $\times$ (so as to write things like $\max_{d_3}((P_{r_1} \cdot P_{r_2|r_1} \cdot U_{d_2,d_3}) + (P_{r_1} \cdot P_{r_2|r_1} \cdot U_{r_2,d_1,d_3})) = P_{r_1} \cdot P_{r_2|r_1} \cdot \max_{d_3}(U_{d_2,d_3} + U_{r_2,d_1,d_3}))$. Then, $D_\Sigma$ can be used for $\sum_{r_1}$, and $D_{\max}$ can be used for $\max_{d_2}$. After those steps, the recomposition rule $R_{\max}$, which uses the monotonicities of $\times$ and $+$, reveals that the elimination order between $d_2$ and $d_3$ is actually free. This was not obvious from the initial $Sov$ sequence. The approach using potentials is unable to make such freedoms explicit, which may induce exponential increase in complexity as shown in 2.3.

**Rule application order**  A chaotic iteration of the rules does not converge, since e.g., rules $D_{\max}$ and $R_{\max}$ may be infinitely alternately applied. Hence, we specify an order in which we apply rules to converge to a unique final DAG of computation nodes (we have used this order in the previous examples). We successively consider each operator-variable pair of the initial sequence $Sov_0$ from the right to the left (marginalizations like $\sum_{x_1,\ldots,x_n}$ can be split into $\sum_{x_1} \cdots \sum_{x_n}$).

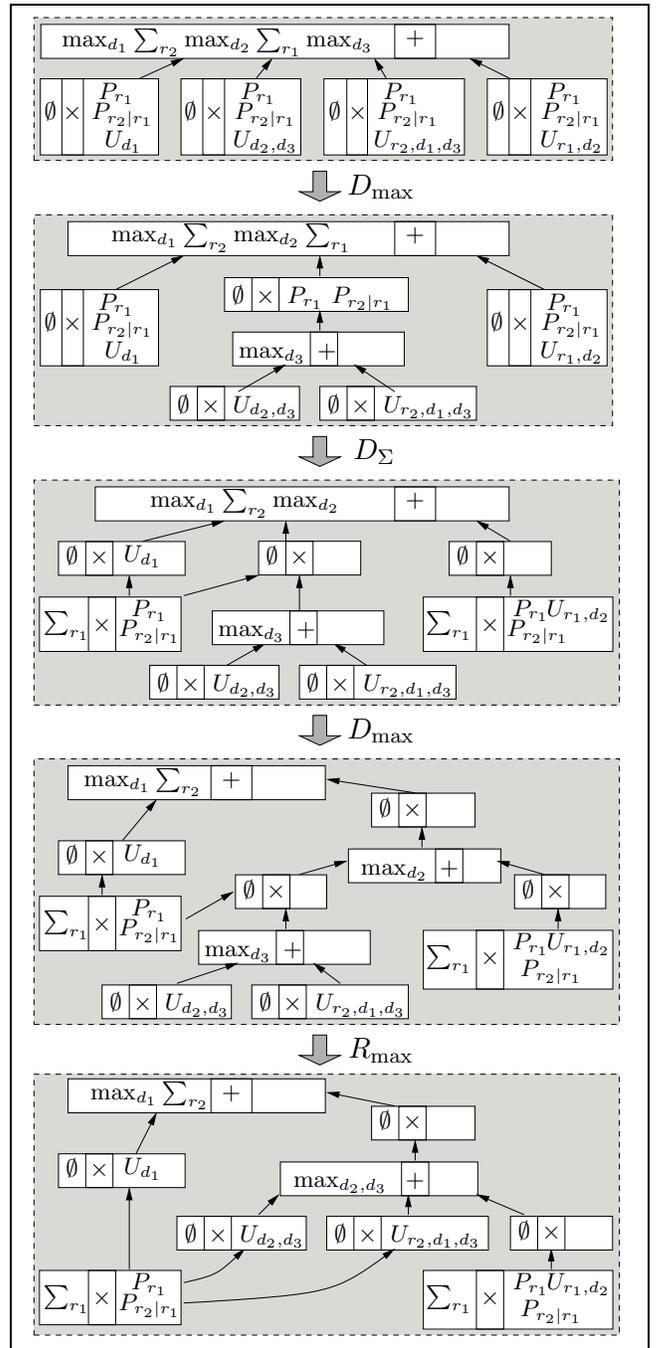

**Figure 3:** Application of rewriting rules for max (the application of the rules may create nodes looking like $(\emptyset, \times, \{n\})$, which perform no computations; these nodes can be eliminated at a final step).

If the rightmost marginalization in the $Sov$ sequence of the root node is $\sum_x$, then rule $D_\Sigma$ is applied once. It creates new grandchildren nodes for the root, for each of which, we try to apply rule $R_\Sigma$ in order to reveal freedoms in the elimination order. If $R_\Sigma$ is applied, this creates new computation nodes, on each of which simplification rules $S_\Sigma^1$ and then $S_\Sigma^2$ are applied (until they cannot be applied anymore).

If the rightmost marginalization in the *Sov* sequence of the root node is $\max_x$, then rule $D_{\max}$ is applied once. This creates a new child and a new grandchild for the root. For the created grandchild, we try to weaken constraints on the elimination order using $R_{\max}$.

Therefore, the rewriting rules are applied in a deterministic order, except from the freedom left when choosing the next variable in $S$ to consider for marginalizations like $\sum_S$ or $\max_S$. It can be shown that this freedom does not change the final structure. The soundness of the macrostructure obtained is provided by the soundness of the rewriting rules:

**Proposition 4.** *Rewriting rules $D_\Sigma$, $R_\Sigma$, $S_\Sigma^1$, $S_\Sigma^2$, $D_{\max}$ and $R_{\max}$ are sound, i.e. for any of these rules $n_1 \rightsquigarrow n_2$, if the preconditions are satisfied, then $val(n_1) = val(n_2)$ holds. Moreover, rules $D_{\max}$ and $R_{\max}$ leave the set of optimal decision rules unchanged.*

**Complexity issues** An architecture is usable only if it is reasonable to build it. Proposition 5 makes it possible to save some tests during the application of the rewriting rules, and Proposition 6 gives upper bounds on the complexity.

**Proposition 5.** *Except for $S_\Sigma^1$, the preconditions of the rewriting rules are always satisfied.*

**Proposition 6.** *The time and space complexity of the application of the rewriting rules are $O(|C \cup D| \cdot |U| \cdot (1+|P|)^2)$ and $O(|C \cup D| \cdot (|U|+|P|))$ respectively.*

### 3.2 TOWARDS MCDAGS

The rewriting rules enable us to transform the initial multi-operator computation node $n_0$ into a DAG of mono-operator computation nodes looking like $(\max_S, +, N)$, $(\sum_S, \times, N)$, $(\emptyset, \times, N)$, or $\phi \in P \cup U$. For nodes $(\max_S, +, N)$ or $(\sum_S, \times, N)$, it is time to use freedoms in the elimination order. To do so, usual junction tree construction techniques can be used, since on one hand, $(\mathbb{R}, \max, +)$ and $(\mathbb{R}, +, \times)$ are commutative semirings, and since on the other hand, there are no constraints on the elimination order inside each of these nodes (the only slight difference with usual junction trees is that only a subset of the variables involved in a computation node may have to be eliminated, but it is quite easy to cope with this).

To obtain a good decomposition for nodes $n$ like $(\max_S, +, N)$ or $(\sum_S, \times, N)$, one can build a junction tree to eliminate $S$ from the hypergraph $G = (sc(N), \{sc(n') \,|\, n' \in N\})$. The optimal induced-width which can be obtained for $n$ is $w(n) = w_{G,S}$, the induced-width of $G$ for the elimination of the variables in $S$.[3] The induced-width of the MCDAG architecture is then defined by $w_{mcdag} = \max_{n \in \mathfrak{N}} w(n)$, where $\mathfrak{N}$ is the set of nodes looking like $(\max_S, +, N)$ or $(\sum_S, \times, N)$.

After the decomposition of each mono-operator computation node, one obtains a *Multi-operator Cluster DAG*.

**Definition 3.** *A* Multi-operator Cluster DAG *is a DAG where every vertex $c$ (called a cluster) is labeled with four elements: a set of variables $V(c)$, a set of scoped functions $\Psi(c)$ taking values in a set $E$, a set of son clusters $Sons(c)$, and a couple $(\oplus_c, \otimes_c)$ of operators on $E$ s.t. $(E, \oplus_c, \otimes_c)$ is a commutative semiring.*

**Definition 4.** *The value of a cluster $c$ of a MCDAG is given by $val(c) = \oplus_{cV(c)-V(pa(c))} \left( \left( \otimes_{c\psi \in \Psi(c)} \psi \right) \otimes_c \left( \otimes_{cs \in Sons(c)} val(s) \right) \right)$. The value of a MCDAG is the value of its root node.*

Thanks to Proposition 7, working on MCDAGs is sufficient to solve influence diagrams.

**Proposition 7.** *The value of the MCDAG obtained after having decomposed the macrostructure is equal to the maximal expected utility. Moreover, for any decision variable $D_k$, the set of optimal decision rules for $D_k$ in the influence diagram is equal to the set of optimal decision rules for $D_k$ in the MCDAG.*

### 3.3 MERGING SOME COMPUTATIONS

There may exist MCDAG clusters performing exactly the same computations, even if the computation nodes they come from are distinct. For instance, a computation node $n_1 = (\sum_{x,y}, \times, \{P_x, P_{y|x}, U_{y,z}\})$ may be decomposed into clusters $c_1 = (\{x\}, \{P_x, P_{y|x}\}, \emptyset, (+, \times))$ and $c_1' = (\{y\}, \{U_{y,z}\}, \{c_1'\}, (+, \times))$. A computation node $n_2 = (\sum_{x,y}, \times, \{P_x, P_{y|x}, U_{y,t}\})$ may be decomposed into clusters $c_2 = (\{x\}, \{P_x, P_{y|x}\}, \emptyset, (+, \times))$ and $c_2' = (\{y\}, \{U_{y,t}\}, \{c_2'\}, (+, \times))$. As $c_1 = c_2$, some computations can be saved by merging clusters $c_1$ and $c_2$ in the MCDAG. Detecting common clusters is not as easy as detecting common computation nodes.

---
[3] For $(\max_S, +, N)$ nodes, which actually always look like $(\max_S, +, \{(\emptyset, \times, N'), N' \in \mathfrak{N}\})$, better decompositions can be obtained by using another hypergraph. In fact, for each $N' \in \mathfrak{N}$, there exists a unique $n \in N'$, denoted $N'[u]$, s.t. $n$ or its children involve a utility function. It is then better to consider the hypergraph $(sc(N), \{sc(N'[u]) \,|\, N' \in \mathfrak{N}\})$. This enables to figure out that e.g. only two variables ($x$ and $y$) must be considered if one eliminates $x$ first in a node like $(\max_{xy}, +, N) = (\max_{xy}, +, \{(\emptyset, \times, U_{y,z}), (\emptyset, \times, \{n_z, U_{x,y}\}), (\emptyset, \times, \{n_z, U_x\})\})$, since $n_z$ is a factor of both $U_{x,y}$ and $U_x$. We do not further develop this technical point.

To sum up, there are three steps to build the architecture. First, the initial multi-operator computation node is transformed into a DAG of mono-operator computation nodes (via sound rewriting rules). Then, each computation node is decomposed with a usual junction tree construction. It provides us with a MCDAG, in which some clusters can finally be merged.

## 4 VE ALGORITHM ON MCDAGs

Defining a VE algorithm on a MCDAG is simple. The only difference with existing VE algorithms is the multi-operator aspect for both the marginalization and the combination operators used. As in usual architectures, nodes send messages to their parents. Whenever a node $c$ has received all messages $val(s)$ from its children, $c$ can compute its own value $val(c) = \oplus_{cV(c)-V(pa(c))} \left( \left( \otimes_{c\psi \in \Psi(c)} \psi \right) \otimes_c \left( \otimes_{cs \in Sons(c)} val(s) \right) \right)$ and send it to its parents. As a result, messages go from leaves to root, and the value computed by the root is the maximal expected utility.

## 5 COMPARISON WITH EXISTING ARCHITECTURES

Compared to existing architectures on influence diagrams, MCDAGs can be exponentially more efficient by strongly decreasing the constrained induced-width (cf Section 2.3), thanks to (1) the duplication technique, (2) the analysis of extra reordering freedoms, and (3) the use of normalizations conditions. One can compare these three points with existing works:

- The idea behind duplication is to use all the decompositions (independences) available in influence diagrams. An influence diagram actually expresses independences on one hand on the global probability distribution $P_{C|D}$, and on the other hand on the global utility function. MCDAGs separately use these two kinds of independences, whereas a potential-based approach uses a kind of weaker "mixed" independence relation. Using the duplication mechanism during the MCDAG building is better, in terms of induced-width, than using it "on the fly" as in [Dechter, 2000].[4]

- Weakening constraints on the elimination order can be linked with the usual notion of *relevant information* for decision variables. With MCDAGs, this notion is not used only for decision rules conciseness reasons: it is also used to reveal reordering freedoms, which can decrease the time complexity. Note that some of the ordering freedom here is obtained by synergism with the duplication.

---

[4]E.g., for the quite simple influence diagram introduced in Section 3.1.1, the algorithm in [Dechter, 2000] gives 2 as an induced-width, whereas MCDAGs give an induced-width 1. The reason is that MCDAGs allow to eliminate both $x_1$ before $x_2$ in the subproblem corresponding to $U_{d,x_2}$ and $x_2$ before $x_1$ in the subproblem corresponding to $U_{d,x_1}$.

- Thanks to simplification rule $S_\Sigma^1$, the normalization conditions enable us not only to avoid useless computations, but also to improve the architecture structure ($S_\Sigma^1$ may indirectly weaken some constraints on the elimination order). This is stronger than Lazy Propagation architectures [Madsen and Jensen, 1999], which use the first point only, during the message passing phase. Note that with MCDAGs, once the DAG of computation nodes is built, there are no remaining normalization conditions to be used.

Compared to existing architectures, MCDAGs actually always produce the best decomposition in terms of constrained induced-width, as Theorem 1 shows.

**Theorem 1.** *Let $w_{G_p}(\preceq_p)$ be the constrained induced-width associated with the potential-based approach (cf Section 2.2). Let $w_{mcdag}$ be the induced-width associated with the MCDAG (cf Section 3.2). Then, $w_{mcdag} \leq w_{G_p}(\preceq_p)$.*

Last, the MCDAG architecture contradicts a common belief that using *division operations* is necessary to solve influence diagrams with VE algorithms.

## 6 POSSIBLE EXTENSIONS

The MCDAG architecture has actually been developed in a generic algebraic framework which subsumes influence diagrams. This framework, called the Plausibility-Feasibility-Utility networks (PFUs) framework [Pralet *et al.*, 2006a], is a generic framework for sequential decision making with possibly uncertainties (plausibility part), asymmetries in the decision process (feasibility part), and utilities. PFUs subsume formalisms from quantified boolean formulas or Bayesian networks to stochastic constraint satisfaction problems, and even define new frameworks like possibilistic influence diagrams. This subsumption is possible because the questions raised in many existing formalisms often reduce to a sequence of marginalizations on a combination of scoped functions. Such sequences, a particular case of which is Equation 1, can be structured using rewriting rules as the ones previously presented, which actively exploit the algebraic properties of the operators at stake.

Thanks to the generic nature of PFUs, extending the previous work to a possibilistic version of influence diagrams is trivial. If one uses the possibilistic pessimistic expected utility [Dubois and Prade, 1995], the optimal utility can be defined by (the probability distributions $P_i$ become possibility distributions, and the utilities $U_i$ become preference degrees in $[0, 1]$):

$$\min_{I_0} \max_{D_1} \ldots \min_{I_{q-1}} \max_{D_q} \min_{I_q} \max \left( \max_{P_i \in P} (1 - P_i), \min_{U_i \in U} U \right).$$

These eliminations can be structured via a MCDAG. The only difference in the rewriting rules is that $\times$ becomes max and $+$ becomes min. The computation nodes then look like $(\min, \max, N)$, $(\max, \min, N)$, or $(\emptyset, \max, N)$, and the MCDAG clusters use $(\oplus_c, \otimes_c) = (\min, \max), (\max, \min)$, or $(\emptyset, \max)$.

# 7 CONCLUSION

To solve influence diagrams, using potentials allows one to reuse existing VE schemes, but may be exponentially sub-optimal. The key point is that taking advantage of the composite nature of graphical models such as influence diagrams, and namely of the algebraic properties of the elimination and combination operators at stake, is essential to obtain an efficient architecture for local computations. The direct handling of several elimination and combination operators in a kind of composite architecture is the key mechanism which allows MCDAGs to always produce the best constrained induced-width when compared to potential-based schemes.

The authors are currently working to obtain experimental results on MCDAGs in the context of the PFU framework (the construction of MCDAG architectures is currently implemented). Future directions could be first to adapt the MCDAG architecture to the case of Limited Memory Influence Diagrams (LIMIDs) [Lauritzen and Nilsson, 2001], and then to use the MCDAG architecture in the context of approximate resolution.

**Acknowledgments**

We would like to thank the reviewers of this article for their helpful comments on related works. This work was partially conducted within the EU IP COGNIRON ("The Cognitive Companion") funded by the European Commission Division FP6-IST Future and Emerging Technologies under Contract FP6-002020.